\documentclass[utf8]{FrontiersinHarvard} 

\usepackage{url,hyperref,lineno,microtype,subcaption}
\usepackage[onehalfspacing]{setspace}

\usepackage{float}
\usepackage{algorithm}
\usepackage{algpseudocode}

\usepackage{caption}
\usepackage{subcaption}

\usepackage{pifont}
\newcommand{\cmark}{\ding{51}}%
\newcommand{\xmark}{\ding{55}}%

\def\keyFont{\fontsize{8}{11}\helveticabold }
\def\firstAuthorLast{Tiago de Souza Farias and Jonas Maziero} 
\def\Authors{Tiago de Souza Farias\,$^{*}$ and Jonas Maziero}
\def\Address{Physics Departament, Center for Natural and Exact Sciences, Federal University of Santa Maria, Roraima Avenue 1000, 97105-900, Santa Maria, RS, Brazil}
\def\corrAuthor{Tiago de Souza Farias}

\def\corrEmail{tiago939@gmail.com}

\begin{document}
\onecolumn
\firstpage{1}

\title[Feature Alignment as a Generative Process]{Feature Alignment as a Generative Process} 

\author[\firstAuthorLast ]{\Authors} 
\address{} 
\correspondance{} 

\extraAuth{}

\maketitle

\begin{abstract}

\section{}
Reversibility in artificial neural networks allows us to retrieve the input given an output. We present feature alignment, a method for approximating reversibility in arbitrary neural networks.
We train a network by minimizing the distance between the output of a data point and the random output with respect to a random input.
We applied the technique to the MNIST, CIFAR-10, CelebA and STL-10 image datasets. We demonstrate that this method can roughly recover images from just their latent representation without the need of a decoder. By utilizing the formulation of variational autoencoders, we demonstrate that it is possible to produce new images that are statistically comparable to the training data. Furthermore, we demonstrate that the quality of the images can be improved by coupling a generator and a discriminator together.
In addition, we show how this method, with a few minor modifications, can be used to train networks locally, which has the potential to save computational memory resources.

\tiny
 \keyFont{ \section{Keywords:} {machine learning; neural network; generative process; reversibility; local training} } 
\end{abstract}

\section{Introduction}

Feature visualization~\cite{olah_feature_2017} is a set of techniques for neural networks aiming to find inputs that maximize the activation of one or more selected neurons from the same network. Usually, feature visualization is used as a method for model interpretability, where one seeks to understand a neural network by analyzing how much each neuron contributes to a neural network by perceiving the images generated by these techniques. The~process of obtaining these inputs is, in~a sense, an~attempt towards reversing a neural network. Since a neural network is composed by functions that map inputs to outputs, the~visual representation of a feature is the input we would have given a target activation for a group of posterior selected~neurons.

The reversibility of neural networks relates to how well one can reverse the map from the activation of target neurons back to the input neurons~\cite{gomez}. In~most cases, neural networks are not reversible, primarily due to three reasons: (1) the presence of non-reversible activation functions (e.g., ReLU~\cite{agarap_deep_2019}), which means that in general, it is impossible to directly recover the input value $x$ given the output value $f(x)$. (2) Non-orthogonal weights, as~there are neither constraints nor incentives for their  matrix representation to converge to having this property. (3) Lack of one-to-one relationships as a result to the reduction of dimension as the information is passed through each layer of a network. In~addition to the reversing mapping, reversible neural networks have the benefit of memory efficiency: unlike non-invertible neural networks, which must store all of the activations for the backward pass during training, reversible neural networks only need to store a portion of the activations in order to update the trainable~parameters. 

Reversibility also constrains the number of possible models, as~many possible parameters configurations model the data. For~example, if~one considers an analytical function that one wants a sufficiently parameterized neural network to approximate, with~the pair of data $\{x, f(x)\}$, several local minima estimate the function $x \rightarrow f(x)$, each of which was obtained by a different random initialization of the neural network parameters (assuming optimal convergence). By~restricting the reversibility $f(x) \rightarrow x$, we can reduce the number of optimal points toward which a neural network can converge. Since many local optima converge to comparable losses, local optima do not pose a problem for neural networks; however, they lack interpretability because the inputs cannot be recovered from a given~output. 

Memory is often a bottleneck for neural networks. Modern deep learning techniques frequently use the backpropagation algorithm~\cite{linnainmaa_taylor_1976, rumelhart_learning_1986}, which requires the storage of all network activations in order to update its parameters. Local training rules enable a more effective memory optimization of neural networks~\cite{baldi_theory_2016}. By~constraining the trainable parameters, such as the weights, to~be updated only by local variables (the information contained in the neurons that share the same parameter), we can reduce the memory requirements to load a model in hardware such as CPUs and GPUs. 
This constraint can conserve memory resources and has a wide range of potential applications, including low-memory devices~\cite{velichko_neural_2020, sohoni_low-memory_2019}, training large batch sizes~\cite{gao_study_2020, you_large_2017}, and, even training very large neural networks~\cite{jing_survey_2019}.

Our goal in this paper is to show that feature alignment can be used for approximate reversibility of neural networks as well as sampling of images. This approximation is based on performing gradient descent on the input space while simultaneously training a network to estimate the input given an output. To~show the feasibility of the proposed technique, we make use  of generative networks to generate samples statistically similar to the training data by making use of approximated reversibility. We also adapt the technique for local training, showing that is possible to reverse an encoder by mapping the output latent vector back to the images of a dataset with only local~variables.

\section{Related Work}

Several works have been done in the area of feature extraction, especially applied for model interpretability and explainability~\cite{shahroudnejad_survey_2021, fan_interpretability_2021, gilpin_explaining_2019, thakur_study_2021, ismail2021improving}. These techniques, used for extracting features, usually consist in activation maximization \cite{mahendran_visualizing_2016, act_max}, where a group of neurons, which can involve from a single neuron up to an entire layer (or channel for convolutions), is selected to extract the feature by maximizing its activation.
Many of these techniques of feature extraction consist in studying features in already pre-trained classifiers~\cite{nguyen_multifaceted_2016}. Other techniques consist in searching for features in the latent space~\cite{shen}. Feature extraction can also be utilized for understanding which parts of an input contribute the most for the target~activations~\cite{selvaraju_grad-cam:_2020, springenberg_striving_2015, zintgraf_new_2017}.

In a generative process, we want to produce new examples with the same statistical distribution as the training data. There are several different techniques to model the data for a generation. Among~these techniques, autoencoder based networks, generative adversarial networks, and~normalizing flows are very popular. Autoencoders (AE), while not generative networks, they constitute of building blocks for other generative networks and offer insights about  mapping the input to other representations. Autoencoder consist of two networks: an encoder that projects the inputs into a vector, usually with a smaller dimension, and~a decoder that reconstructs the input from this vector. The~compressed vector has a high-level representation of the model, in~which each neuron contributes to properties beyond the data level at the input layer \cite{lee_unsupervised_2011}. Autoencoders are commonly trained in an unsupervised fashion, nevertheless, some variants include labeled information to further increase training for a specific objective. Variational autoencoders (VAE) \cite{kingma_auto-encoding_2014, kingma_introduction_2019, doersch_tutorial_2021} gives autoencoders generative capability by projecting the data into a probabilistic latent vector, thus we can generate data statistically similar to the training data by sampling random latent vectors and projecting them to a decoder network.
Generative adversarial networks (GANs) \cite{goodfellow_generative_2014, salehi_generative_2020, gui_review_2020} are another example of a generative method. By~having two competing networks, a~generative network which takes a random low-dimensional input and outputs an image, and~a discriminator network that compares the images from the training dataset and the sampled ones from the generator. The~competition arises by training the generator to fool the discriminator by generating images as closest to the training dataset as possible. Normalizing flows~\cite{papamakarios_neural_2019, kobyzev_normalizing_2020, kingma_glow:_2018} is another generative paradigm that generate images by transforming a simple distribution to a more complex one by a series of reversible~transformations.

Diffusion models is another method that can be utilized to generate samples that are statically comparable to a dataset \cite{diffusion1, diffusion2, diffusion3}. They are trained to predict the noise that is presented in a sample by repeatedly exposing them to increasing levels of noise as part of their training. The state-of-the-art capability of this method, which can generate samples with high fidelity and that are similar to the training dataset, is the primary benefit of using this method. On the other hand, diffusion models take a long time to sample because they require a large number of steps to denoise an image, which is a process that is very computationally intensive. 

There have been works combining autoencoders with GANs \cite{larsen_autoencoding_2016}. The~work done in Refs.~\cite{nguyen_synthesizing_2016,dosovitskiy_generating_2016} is related to ours. They synthesized new images with the same statistics as the training data by inputting features to a generator network. The~main difference is that, in~these previous articles, the~features are obtained with a pretrained~network.

Most works on reversibility consist in architectural changes of neural networks \cite{atapattu_improving_2019, schirrmeister_training_2018, grathwohl_ffjord:_2018, behrmann_invertible_2019, baird_one-step_2005}. These changes guarantee a one-to-one relationship between inputs and outputs. BiGAN \cite{donahue_adversarial_2017} constructs a generative network and a reverse network that inputs images back to noise, which can be used to obtain a latent representation of a dataset directly. Ref.~\cite{dong_deep_2021} shows that is possible to reverse neural networks in the case of reconstruction of~images.

Local learning rules have been explored since Donald Hebb proposed a simple model for learning in the brain~\cite{noauthor_d.o._1999}. The~main advantage of this kind of learning algorithm is requiring lower memory resources. Some works are biologically inspired~\cite{krotov_unsupervised_2019, lindsey_learning_2020}, while others focus solely on efficiency~\cite{isomura_local_2016, isomura_error-gated_2018, wang2021revisiting, guo_backlink:_2022}. There is a growing body of work discussing whether the brain does backpropagation~\cite{whittington_theories_2019, song_can_2020}, with~some approximations for training artificial neural networks~\cite{millidge_activation_2020, salvatori_predictive_2021, lillicrap_backpropagation_2020, laskin2021parallel}.

Another approach for saving memory resources is gradient checkpoint~\cite{checkpoint1, checkpoint2, checkpoint3, checkpoint4}, where memory is traded with computation time by re-evaluating neurons when they are needed for backpropagation instead of storing their activations all at once. While this technique decreases the amount of memory necessary to train a neural network, it requires many forward propagation calculations on the network, depending on its size, which can increase time consumption, while local learning rules, as~opposite, require only one forward propagation to update the~parameters.

Table \ref{table:related_works} summarizes the related works. We classified each method by four properties: being able to train with new data, whether the method can reconstruct data from a latent space, if the method is able to produce new samples, and reversibility. Three methods have all four properties: normalizing flows, diffusion models and feature alignment. Architecturally, normalizing flows is composed entirely by reversible layers, while feature alignment allows for arbitrary networks. Diffusion models contain a reversibility restriction that is reversing samples from noise; consequently, diffusion can only reverse samples with higher noise to lower noise. This requirement does not apply to feature alignment because the mapping can be done with any number of dimensions from the input space to the output space and loss does not always require noise for optimization. 

\begin{table}[h!]
\begin{center}
\begin{tabular}{ | >{\centering\arraybackslash} m{5em} | >{\centering\arraybackslash} m{5em} | >{\centering\arraybackslash} m{7em} | >{\centering\arraybackslash} m{5em} |  >{\centering\arraybackslash} m{5em} | >{\centering\arraybackslash} m{5em} |}
\hline
Method & trainable & reconstruction & sampling & reversible & reference\\ 
\hline
AE & \cmark & \cmark & \xmark & \xmark & \cite{lee_unsupervised_2011} \\
\hline
VAE & \cmark & \cmark & \cmark & \xmark & \cite{kingma_auto-encoding_2014} \\
\hline
GAN & \cmark & \xmark & \cmark & \xmark &  \cite{goodfellow_generative_2014}\\
\hline
BiGAN & \cmark & \cmark & \cmark & \xmark & \cite{donahue_adversarial_2017}\\
\hline
feature extraction & \xmark & \xmark & \xmark & \cmark & \cite{mahendran_visualizing_2016}\\
\hline
normalizing flow & \cmark & \cmark & \cmark & \cmark & \cite{kingma_glow:_2018}\\
\hline
VAE-GAN & \cmark & \cmark & \cmark & \xmark & \cite{larsen_autoencoding_2016}\\
\hline
reversible guidance & \xmark & \cmark & \xmark & \cmark & \cite{atapattu_improving_2019}\\
\hline
diffusion models & \cmark & \cmark & \cmark & \cmark & \cite{diffusion3}\\
\hline
\textbf{feature alignment} & \cmark & \cmark & \cmark & \cmark & \textbf{this work} \\
\hline
\end{tabular}
\caption{Comparative table among previous  works from the literature and with this article.}
\label{table:related_works}
\end{center}
\end{table}
\unskip

\section{Methods}

The method of feature alignment is covered in this section. It consists of two phases: first, we perform a gradient descent on a random input using a loss function that measures the distance between the encoded random input and an encoded image.  After~that, the~network is then trained on a different loss function, which evaluates the distance between the inputs and the gradient that was performed on it. By~doing this,the network is able to learn the inverse map that leads from its outputs to the inputs that correspond to those outputs, thus recovering the information that triggered its~activation.

\subsection{Feature Alignment (FA)}

The feature alignment encoder consists of a encoder with parameters $\theta$ and an arbitrary number of latent variables as the output. From~a dataset $\textbf{x}$
 $\in \textbf{X}$, $\textbf{z}_x = E(\textbf{x};\theta)$ is the output from an input $\textbf{x}$. With~the same network, $z_r = E(\textbf{r};\theta)$ is the output from a random input $\mathbf{r}$, chosen from some probability distribution, with~the same dimension as the input~data.

The feature $\hat{\textbf{r}}$ of $\textbf{z}_x$ is obtained by minimizing a distance function $\mathcal{L}(\textbf{z}_x,\textbf{z}_r)$ with respect to the random inputs $\textbf{r}$. We choose a gradient flow for minimizing this distance, since it can evolve the random input continually, as~follows in the Equation~(\ref{equ:flow}).

\begin{equation}\label{equ:flow}
    \frac{\partial r}{\partial t} = -\frac{\partial \mathcal{L}}{\partial r}.
\end{equation}

Since $\textbf{z}_x$ is fixed, $\textbf{r}$ will evolve such as the function of the random variables will approximate $\textbf{z}(\textbf{x})$ as much as possible (See Appendix~\ref{A.1}). We want to solve Equation~(\ref{equ:flow}) as efficiently as possible in time and memory. By~discretizing the gradient flow, we obtain an approximation for the feature, as~shown in Equation~(\ref{equ:euler}).

\begin{equation}\label{equ:euler}
    r^t = r^{t-1} - \tau \frac{\partial \mathcal{L}}{\partial r},
\end{equation}
with $\tau$ being a hyperparameter that weights the contribution of the gradient. Equation~(\ref{equ:euler}) is similar to activation maximization, except~that we are minimizing for the neurons to have a target activation, which is the latent representation of an image $x$. These updates are done in $T$ time steps. Properly optimized, the~solution to Equation~(\ref{equ:euler}) converges to the input $x$ by approximating the inverse of the weights (see Appendix~\ref{A.1}). So, by~optimizing the parameters of the network, the~weight matrix between layers will have the orthogonal property $\textbf{w}^T \textbf{w} = \textbf{I}$, which implies in approximated reversibility (see Appendix~\ref{A.2}).

\begin{figure}
    \centering
    \includegraphics[scale=0.4]{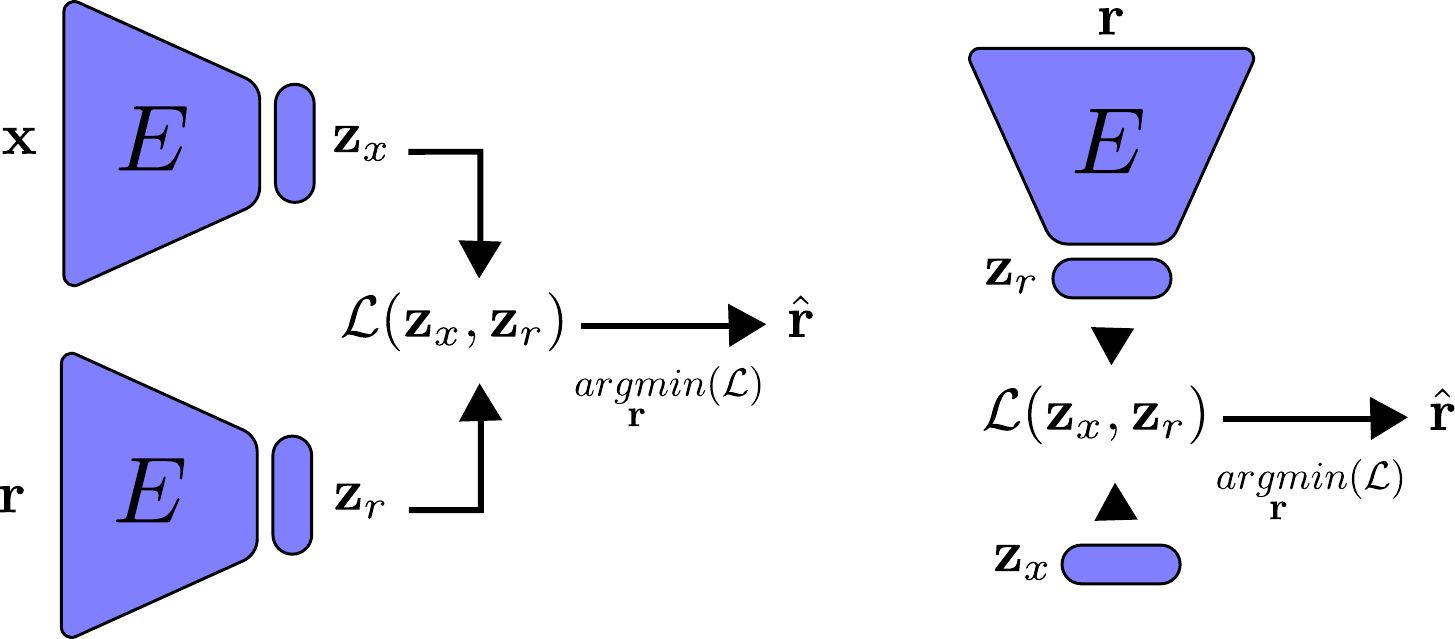}
    \caption{An encoder that has been trained using feature alignment for the purposes of reconstruction. Left: the encoder receives an input that is batched from training data as well as a random input, which is updated by minimizing the distance between their latent representations. The~network is trained to approximate $\hat{\textbf{r}} \approx \textbf{x}$. Right: during inference, we reconstruct $x$ by only using its latent representation $z_r$ and doing gradient descent on the random input $r$.}
    \label{fig:fa}
\end{figure}

After we extract the representation $\hat{\mathbf{r}}$, we measure how similar it is to the inputs $\mathbf{x}$ by a new loss function $\mathcal{C}(\mathbf{x},\hat{\mathbf{r}})$. This second loss function is used for training the encoder by optimizing its parameters. As~the neural network is trained, the~encoder learns, not only to map the inputs to the latent variables, but~also the reconstruction of the inputs from the latent vector. Following training, we can reconstruct the inputs by knowing only the latent vector. Figure~\ref{fig:fa} and Algorithm \ref{algo1} summarize the feature alignment technique. First, using an encoder $E(\theta)$, we obtain the latent representation $z_x$ of an input, $x$, which can be an image. The~same encoder is then given a random input of the same size as the input that is drawn from a predetermined distribution (such as a uniform or Gaussian distribution), outputting the latent representation $z_r$.
We then perform a gradient descent on the random input for $T$ steps  (a chosen hyper-parameter). Once we have both latent representations, we update the encoder parameters $\theta$ by minimizing the distance between the optimized random input $\hat{r}$ and the input $x$. 

\begin{algorithm}[H]
  \caption{Training with feature~alignment}
  \label{algo1}
\begin{algorithmic}
\State $\mathbf{z_x} = E(\mathbf{x}; \theta)$
\State initialize $\mathbf{r} = \mathbf{x}.shape$ from a random distribution
\State $t=0$
\While {$t < T $}
    \State $\mathbf{z_r} = E(\mathbf{r}; \theta)$
    \State $\mathcal{L} = ||\mathbf{z_x} - \mathbf{z_r}||_2^2$
    \State $r = r -\tau \frac{\partial \mathcal{C}}{\partial r}$
    \State $t = t + 1$
\EndWhile
\State $\hat{\textbf{r}} = \textbf{r}$
\State $\mathcal{C} = ||\mathbf{x} - \hat{\textbf{r}}||_2^2$
\State update $\theta$ by optimizing $\mathcal{C}$
\end{algorithmic}
\end{algorithm}
\vspace{-6pt} 

\subsection{A toy example}

To gain a better understanding on how feature alignment works, here we will look at a straightforward example. Let's say we want to approximate a function $y =f(x)$ with a neural network. While we can approximate $y$ with a sufficiently parameterized neural network $\mathcal{F}(x;\theta)$, we can not recover $x$ given only $\mathcal{F}(x;\theta)$ for functions that do not have a one-to-one relationship.

With the feature alignment method, we can constrain the network $\mathcal{F}$ to be able to approximate the reversible map $x = f^{-1}(y)$. In this example, we will look at the function $y=\sin(3\pi x)$. The network used consists of two fully connected hidden layers with 1024 neurons each, with both input and output single neurons. The network trained with FA has an extra neuron in the output layer to act as a latent variable because $\sin(3\pi x)$ has a correspondence many-to-one. The equation
\begin{equation}\label{equ:toy_aux_cost}
    \mathcal{L} = ||l_x - l_r||_2^2 + \alpha ||f_x - f_r||_2^2,
\end{equation}
represents the auxiliary loss function for reversing the network, with a hyper-parameter $\alpha$, which in this example was set to $\alpha=0.01$, $f_x$  is the output of the network we want to approximate and $f_r$ is the output given a random input, $l_x$ is the latent neuron and $l_r$ is the latent neuron from the random input. Then, we can recover $x$ by using the equation $\hat{r} = r - \frac{\partial \mathcal{L}}{\partial r}$.

The parameters of network are updated to minimize both the difference between the input to its approximation and the output to the function we want to approximate, with the loss function:
\begin{equation}
    \mathcal{C} = ||y - f_x||_2^2 + ||x - \hat{r}||_2^2.
\end{equation}
Figure \ref{fig:toy} shows the results for a network trained with and without feature alignment. We can observe that in the absence of FA, the network is only able to approximate the inputs partially.  On the other hand, when the loss of feature alignment is included, we have complete approximation within the entire function domain.

\begin{subfigure}[h]
    \centering
    \setcounter{subfigure}{0}
    \begin{minipage}[b]{0.47\textwidth}
        \includegraphics[width=\linewidth]{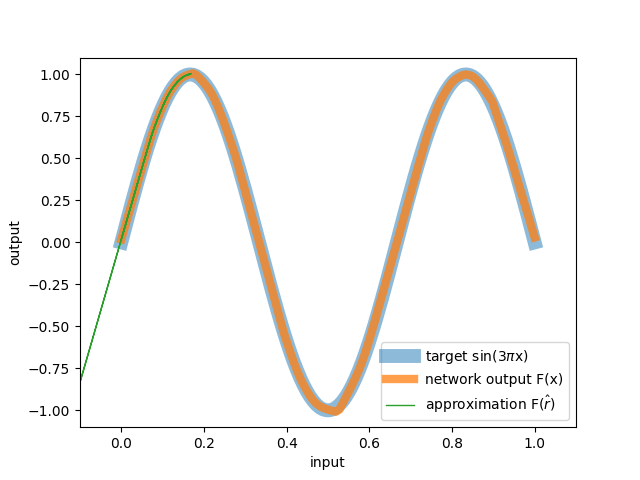}
        \caption{Without feature alignment.}
        \label{fig:toy_notfa}
    \end{minipage}  
   \setcounter{subfigure}{1}
    \begin{minipage}[b]{0.47\textwidth}
        \includegraphics[width=\linewidth]{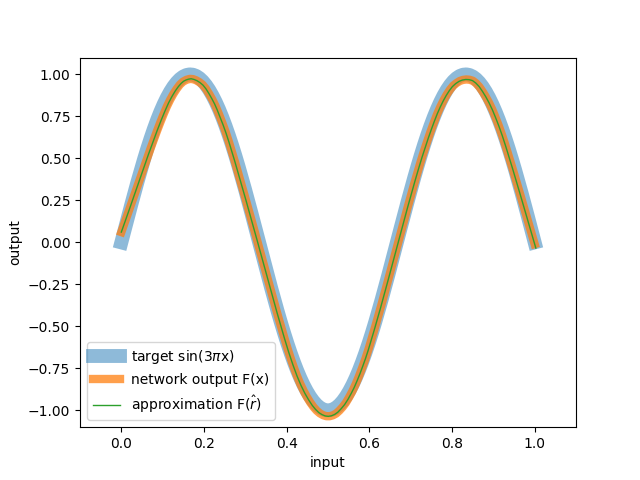}
        \caption{With feature alignment.}
        \label{fig:toy_fa}
    \end{minipage}
    \setcounter{subfigure}{-1}
    \caption{Approximation of reversibility. (\textbf{a}) The network trained without feature alignment can only recover partially the inputs. (\textbf{b}) The network trained with feature alignment can retrieve all the inputs.}
    \label{fig:toy}
\end{subfigure}

\subsection{Variational Autoencoders with Feature Alignment (VFA)}

In the context of generative processes,  autoencoders, in~general, are unable to generate new samples with the same statistical distribution as the training data. The~latent variables from the data, if~associated with a distribution of variables, may be too complicated or convoluted for effective sampling. To~enable feature alignment with sampling, we use the variational autoencoder (VAE) formulation, without~a decoder network. As~a result, the~inverse of the encoder becomes its own decoder, just as was previously with autoencoders. In~the VAE, the~output of the encoder is coupled with two layers that return the mean value $\mu_x$ and variance $\sigma_x^2$ of the data. We constrain the latent vector to have a distribution that is easy to sample (typically a Gaussian distribution), by~comparing two probability distributions using a metric such as the Kullback-Leibler divergence. Subsequently, the~cost function in Equation~(\ref{equ:cost_feature}) is used to train the feature encoder with a constraint to the output latent variables from a known random probability distribution $p(\mathbf{z}),$ from which we can easily sample. The~constant $\beta$ is a hyper-parameter that improves the disentanglement representation of the data by regularizing the latent vector~\cite{higgins_beta-vae:_2016, burgess_understanding_2018, sikka_closer_2019}:
\begin{equation}\label{equ:cost_feature}
    \mathcal{L} = ||\textbf{x} - \hat{\textbf{r}}||_2^2 - \beta \mathcal{D}_{KL}(q_{\theta}(\textbf{z}|\textbf{x})||p(\textbf{z}))
\end{equation}

\begin{figure}[H]
    \centering
    \includegraphics[scale=0.4]{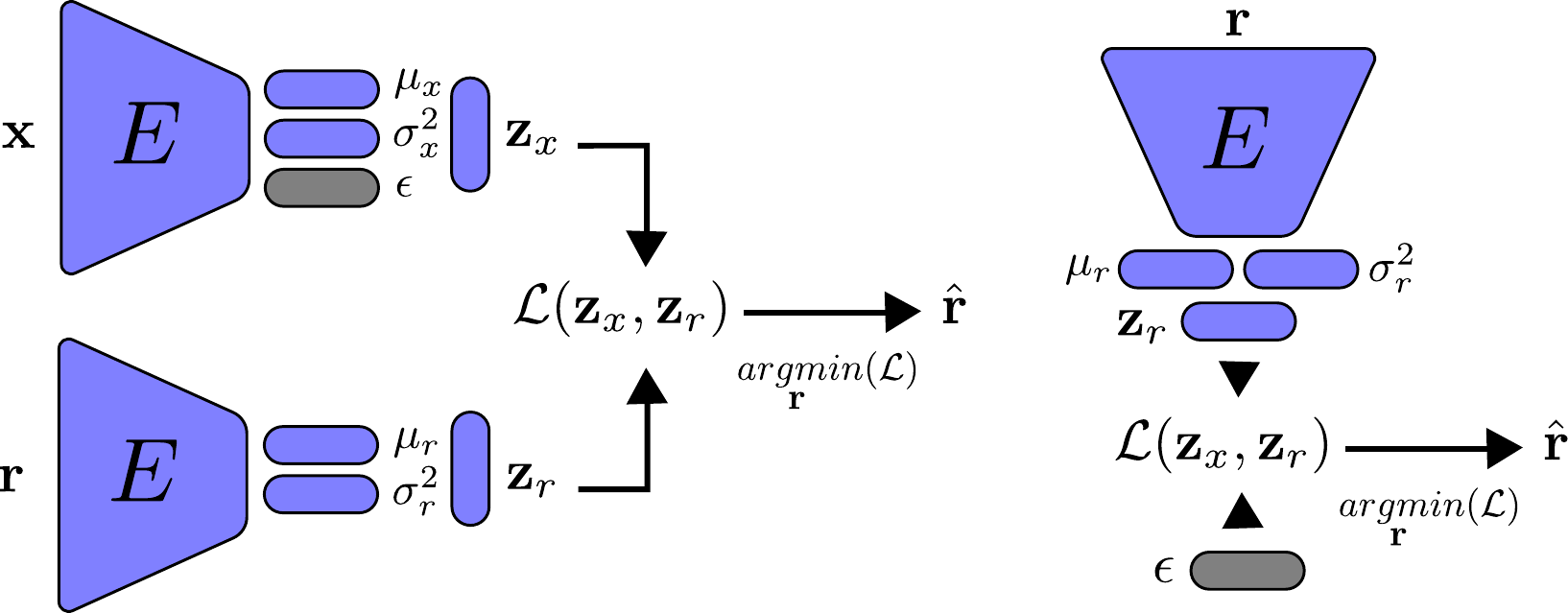}
    \caption{Variational autoencoder with feature alignment. Left: training the encoder to reconstruct the inputs $\mathbf{x}$. Right: we randomly sample a normal vector $\epsilon$ to generate new data statiscally simillar to the training data. Note that the main different between encoders with FA and VFA is on the latent~representation.}
    \label{fig:vae}
\end{figure}

The distribution $p(\mathbf{z})$ is chosen according to the principle of maximum entropy: since the latent variables are in the range $(-\infty, +\infty)$, the~Gaussian distribution is the most appropriate for this case. Each latent variable is then constrained to have a Gaussian distribution with zero mean and one variance. Similar to VAE, we cannot train the encoder by directly sampling the mean and variance of the latent vector. Instead, we employ the re-parametrization trick: we sample a random vector $\epsilon$ from a normal distribution, the~latent vector is represented as $\mathbf{z_x} = \mu_x + \epsilon \odot \sigma^2_x$, with~$\odot$ the element-wise product. Because~the role of $\mathbf{z_r}$ is only to reconstruction, it should be noted that we do not have a random normal vector for this variable. Figure~\ref{fig:vae} summarizes training a VAE with feature~alignment.

\unskip


\subsection{Improving the Quality of the Features (VFA-GAN)}

As will be shown in the results section, the~images extracted using the feature alignment trained with VAE are blurry, due to the variational autoencoder nature~\cite{rezende_taming_2018}.  We add a generator network $G$ and a discriminator network $D$ to the images generated by the technique to improve their quality. In~this manner, the~optimized random vector functions as a second latent vector, thereby sampling a more complex distribution from the latent representation. This generator network is similar to the refiner network described in Ref.~\cite{atapattu_improving_2019}, which takes an image as input and outputs an improved version of~it.

The optimized random input is fed into the generator, which then generates a new output that is compared to the input $\textbf{x}$. The~discriminator is trained as a generative adversarial network, that assesses the likelihood that $G(\mathbf{\hat{r}})$ is genuine or fake (that is, i.e.,~whether it comes from training data or not) . The~generator is updated by receiving gradients from the discriminator. For~more stable training, we use the least square loss for the discriminator~\cite{mao_least_2017}. Alternatively, it may be possible to use the Wasserstein GAN formulation~\cite{arjovsky_wasserstein_2017}, which replaces the discriminator with a critic network that measures the score of the \textit{realness} of an image.

We propose using a random schedule for the variable $\beta$ in order to reduce the potential effects that could be caused by the posterior collapse problem in VAEs~\cite{takida_preventing_2021, havrylov_preventing_2020, lucas_dont_2019} and to maintain a balance with the reconstruction loss. We take a sample from a uniform distribution $\beta \leftarrow \mathcal{U}(0,1)$ for every example that we go through in the training~process.

Although pixel-level loss is typically used to optimize image reconstruction, high-level data properties can also be considered. Perceptual loss~\cite{johnson_perceptual_2016}  is a type of measurement that compares the output of the reconstruction with the original image at high-level neurons (presented near the end of the network).
The mean and variance layers from the encoder network are used in this case as the perceptual loss, requiring the reconstruction to have the same statistical properties as the original~input.

The final losses, for~the encoder, generator, and~discriminator are shown in \linebreak Equations~(\ref{equ:L_E})--(\ref{equ:L_D}) respectively:
\begin{align}
    \mathcal{L}_{E} &= ||\textbf{x} - \hat{\textbf{r}}||_2^2 + \beta \mathcal{D}_{KL}(q_{\theta}(\textbf{z}|\textbf{x})||p(\textbf{z})) \label{equ:L_E}, \\
    \mathcal{L}_{G}& = ||1 - D(G(\hat{\textbf{r}}))||_2^2 + \lambda (||\mu(\textbf{x})-\mu(G(\hat{\textbf{r}}))||_2^2 + ||\sigma^2(\textbf{x})-\sigma^2(G(\hat{\textbf{r}}))||_2^2) \label{equ:L_G}, \\
    \mathcal{L}_{D} &= ||1 - D(\textbf{x})||_2^2 + ||-1 - D(G(\hat{\textbf{r}}))||_2^2 \label{equ:L_D}.
\end{align}
with $\lambda$ a hyperparameter that weights the perceptual loss~contribution.

\begin{figure}[H]
    \centering
    \includegraphics[scale=0.4]{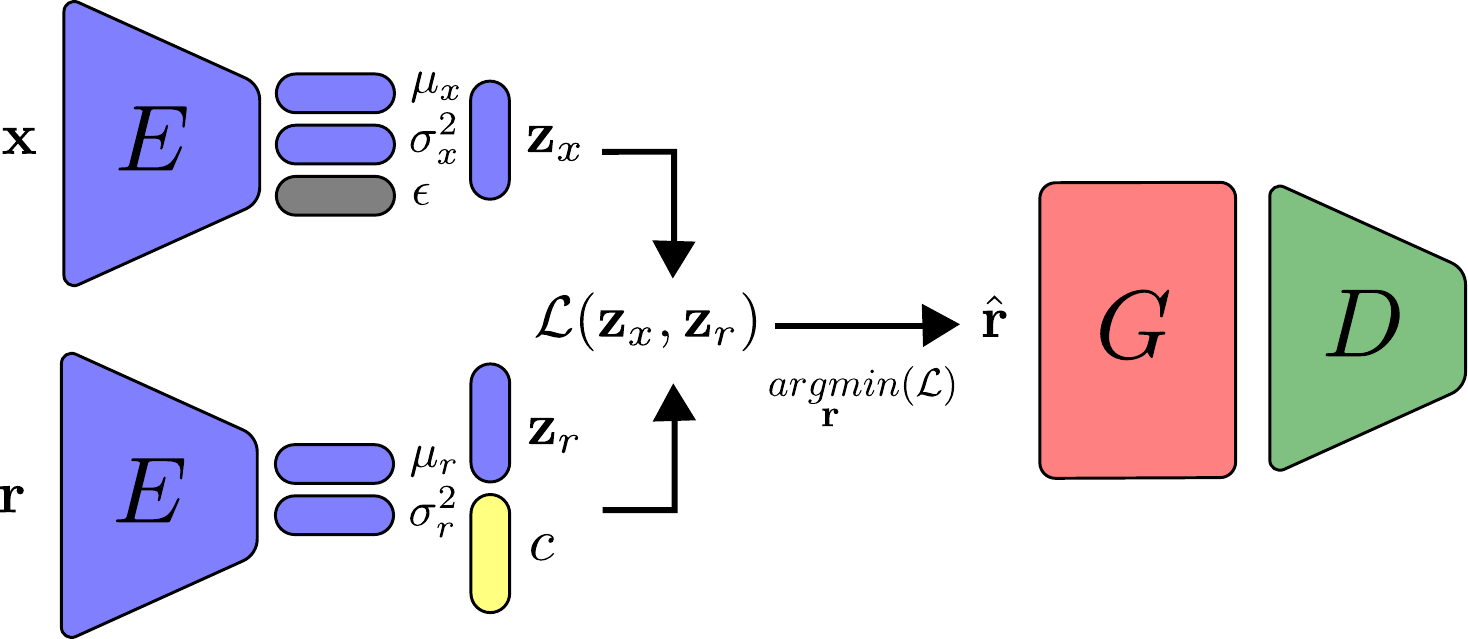}
    \caption{Training in the VFA-GAN setting, with~the addition of generator and discriminator networks.}
    \label{fig:img11}
\end{figure}

If training data contains additional information, such as labels, it is possible to train a neural network simultaneously with supervised training for specific related tasks such as conditional generation, in~which samples must correspond to a desired class. We can condition the latent variables to have different distributions from different classes. The~output of the network is trained using the Gaussian distribution $\textbf{z} \sim Q(z_i,c_i),$ with $c_i$ being a one-hot vector containing the class information. Similar to the work done in Ref.~\cite{ardizzone_analyzing_2019}, we couple a linear classification layer on top of the network, parallel to the mean and variance layers. 
Since the class layer is linear, we can choose a higher value for the one-hot vector during inference time to emphasize the selected class.  The~improvements made on the variational autoencoders with feature alignment discussed in this section are illustrated in Figure~\ref{fig:img11}.

\subsection{Local Feature~Alignment}

\begin{figure}[h]
    \centering
    \includegraphics[scale=0.3]{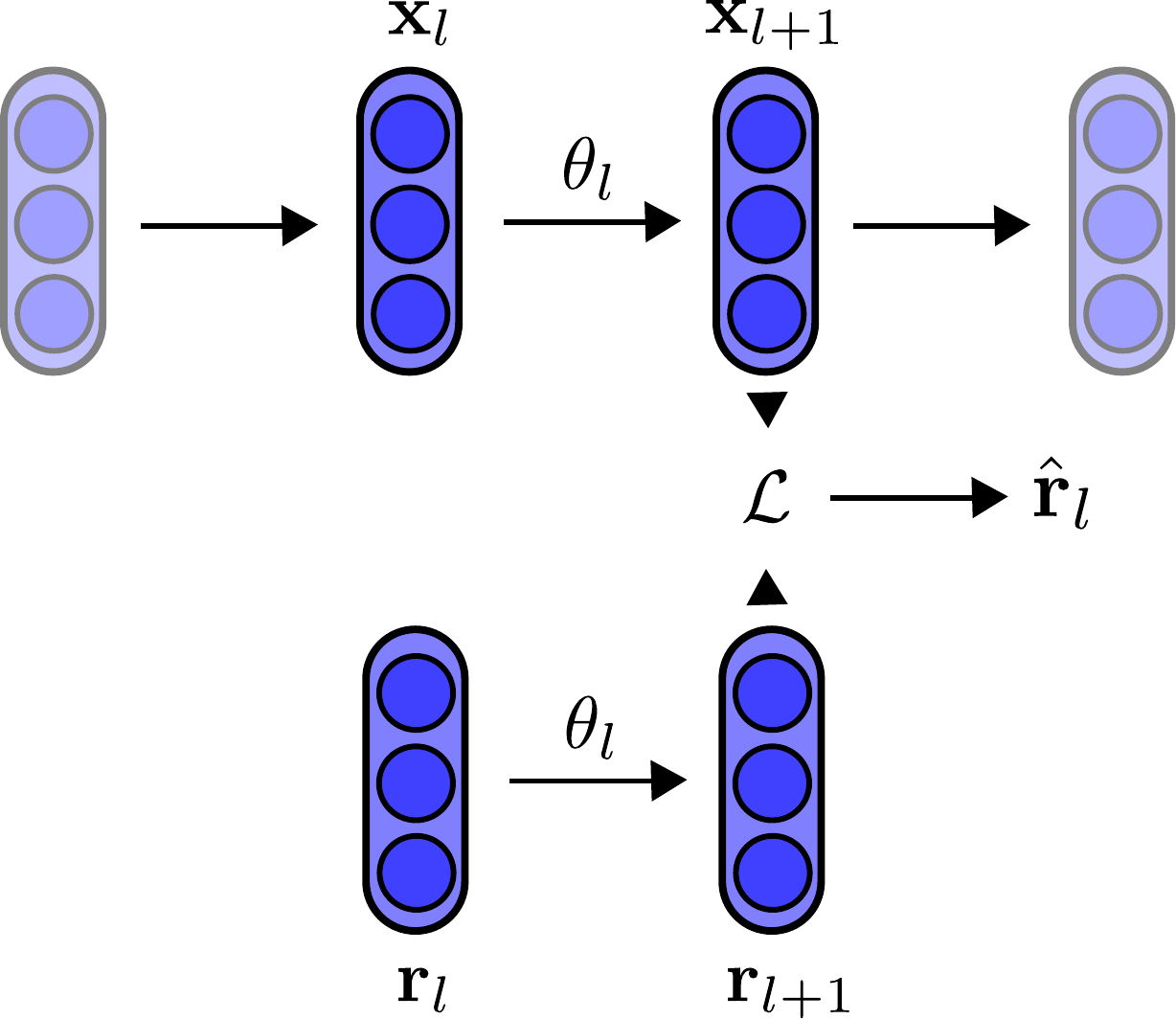}
    \caption{Illustration of the local training rule. Within~the network, we select a pair of layers input-output $l$ and $l+1$. The~parameters $\theta_l$ are updated by minimizing the distance between $x$ and $\hat{r}_l$, in~which is obtained by minimizing the distance $x_{l+1}$ and $r_{l+1}$ with respect to $r_l$.}
    \label{fig:local}
\end{figure}

The rules for feature alignment were presented as a global rule: the auxiliary loss is defined with the output layer (with the pair $z_x$ and $z_r$) and the loss is defined with the input layer (with the pair $x$ and $\hat{r}$), allowing full exchange of information between all layers. However, we can reformulate this rule with local losses, similar to target propagation rules~\cite{farias_gradient_2018, bengio_how_2014, ororbia_conducting_2018}: the auxiliary loss and loss are defined as the interaction between two connected layers only (or even individual neurons), as~follows: for each layer $l$, from~the first to the last, we activate $x_{l+1}$ from its inputs $x_l$ and store a second activation from a random input $r_l$ with the same dimension. We then optimize the random input $r_l$ with an auxiliary loss between activation of the random output $r_{l+1}$ and the true output $x_{l+1}$. Finally, the~parameters of the chosen layer $\theta_l$ are updated by optimizing the loss between the reconstruction $\hat{r}_l$ and true input $x_l$. This technique of local training is summarized in Algorithm \ref{algo2} and illustrated in Figure~\ref{fig:local}.

\begin{algorithm}[H]
  \caption{Training with local feature~alignment}
  \label{algo2}
  \begin{algorithmic}[1]
    \For {l=0, L} \Comment{for each layer}
        \State $\textbf{z}_x^l = E(\textbf{x}; \theta_l)$
        \State \textbf{initialize} $\textbf{r} = \textbf{x}.shape$
        \State $t = 0$
        \While {$t < T $}
            \State $\textbf{z}_r^l = E(\textbf{r}; \theta_l)$
            \State $\mathcal{L}_l = ||\textbf{z}_x^l - \textbf{z}_r^l||_2^2$
            \State $r = r -\tau \frac{\partial \mathcal{L}_l}{\partial r}$
            \State $t = t + 1$
        \EndWhile
        \State $\hat{\textbf{r}} = \textbf{r}$
        \State $\mathcal{C}_l = ||\textbf{x} - \hat{\textbf{r}}||^2$
        \State \textbf{update} $\theta_l$ by optimizing $\mathcal{C}_l$
        \State $\textbf{x} = \textbf{z}_x^l .detach$
    \EndFor
  \end{algorithmic}
\end{algorithm}

Each layer of the neural network trains its parameters to become a predictive machine by attempting to predict the inputs using knowledge of the outputs.
The local learning constraint has a greater impact on the non-linearity of a neural network trained in this manner.
Local rules can only rely on very strict information content available, whereas backpropagation can adjust all network parameters so that the feature reconstructs the input. Non-reversible functions, like the ReLU function, propagate loss of information, resulting in low-fidelity reconstructions.
In order to retain as much information as possible, a~non-linear function must be carefully chosen. The~function \textit{inverse hyperbolic sine} ($ arcsinh(x) = \ln(x + \sqrt{1+x^2}))$), is similar to the \textit{hyperbolic tangent}
 near zero and logarithmic at large (absolute) values. This function has the properties of being fully invertible, zero-centered mean, unbounded, continuously differentiable and its gradient does not vanish as fast as for $tanh$. These properties make $arcsinh$ a good candidate function for local~training.

Similar to how non-local feature alignment is typically done, we must propagate the information backward from the output to the input layer by layer at inference time. However, the~non-linear function will be crucial in this situation because it must be reversible in order to approximate reversibility.
This is accomplished by applying the inverse of the non-linear function after each layer that makes use of the function, following the input of the latent vector. The~Algorithm \ref{algorithm3} provides a summary of this~procedure.

\begin{algorithm}[H]
  \caption{Reconstruction with local feature~alignment}
  \label{algorithm3}
  \begin{algorithmic}[1]
    \State sample $\mathbf{z}_x^L$
    \For {l=L, 0}
      \State \textbf{initialize} $\mathbf{r} = \mathbf{x}.shape$ with $\mathbf{x}$ as $\mathbf{z}_x^l = E(\mathbf{x}; \theta_l)$
      \State $t = 0$
      \While {$t < T $}
        \State $\mathbf{z}_r^l = E(\mathbf{r}; \theta_l)$
        \State $\mathcal{L}_l = ||\mathbf{z}_x^l - \mathbf{z}_r^l||_2^2$
        \State $r = r -\tau \frac{\partial \mathcal{L}_l}{\partial r}$
        \State $t = t + 1$ \
      \EndWhile
      \State $\hat{\mathbf{r}} = \mathbf{r}$
      \If {layer = non-linear function $f$}
        \State $\hat{\mathbf{r}}  = f^{-1}(\hat{\mathbf{r}} )$
      \EndIf
      \State $\mathbf{z}_x^l = \hat{\mathbf{r}} $
    \EndFor
  
  \end{algorithmic}
\end{algorithm}


\section{Implementation~Details}

The encoder network consists of a series of convolutional layers, similar to the AlexNet architecture~\cite{krizhevsky_imagenet_2012}, but~with stride one and two for down-scaling, instead of maxpool, with~LeakyReLU activation. The~generator network has three convolutional layers. The~discriminator network has the same architecture as the encoder, except~for the last layer that outputs a single value. Only the generator utilizes batch normalization after each convolution. All convolutions have kernel size $k=3$. Details of the networks can be found in Tables~\ref{tab:network_mnist}--\ref{tab:network_celeba} for MNIST, CIFAR-10, CelebA and STL-10 respectively.

We use the Adam optimizer~\cite{kingma_adam:_2017} with  learning rate $\eta=0.00001$ and batch size $128$. The~parameters of the encoder and generator networks are initialized with orthogonal initialization~\cite{saxe_exact_2014, hu_provable_2020}. We set the hyperparameter $\lambda=0.01$ and sample $\beta$ from a uniform distribution, with~a different random value for each training example. We set the hyper-parameter $\tau=1$ for the reconstruction of the input at one-shot $T=1$.

From \ref{A.2}, we have that the loss becomes unstable when the weights $w^2 > 2$, so we restrict the weights to the range $-\sqrt{2} \le w \le \sqrt{2}$ by clamping then, as~shown in~Equation~(\ref{equ:clamp}).
\begin{equation}\label{equ:clamp}
    w = 
\begin{cases}
    -\sqrt{2} & \text{if } w < -\sqrt{2}, \\
    \sqrt{2} & \text{if } w > \sqrt{2},\\
    w              & \text{otherwise}.
\end{cases}
\end{equation}

We also report the results of the modified feature alignment for local training as a proof of concept by training an encoder for reconstruction from a latent~vector. For GAN, when used for reconstruction, we search the latent space that leads to most similar images by optimizing $argmin_z ||x - G(z)||_2^2$.

\section{Results}

We compare the results against traditional variational autoencoders and generative adversarial networks. The~results show the reconstruction of the inputs using feature alignment and generator applied to the generated input. Additionaly, we also show random samples from the generator network. Furthermore, we display the model size and inference time for each dataset in the next section. All models were trained and evaluated on an Nvidia RTX 2070 graphics processing unit (GPU) card.

We measured the quality of the results by using the Fr\'echet Inception Distance (FID)~\cite{heusel_gans_2018, Seitzer2020FID}). The~FID score is calculated by extracting the activation of the global spatial pooling layer of a pre-trained Inception V3 model~\cite{szegedy_rethinking_2016}, for~equally numbered images from the dataset (here we choose 10,000 images) and sampled from a generator model, as~shown in~Equation~(\ref{equ:fid}).
\begin{equation}\label{equ:fid}
    FID = ||\mu_1 - \mu_2||_2^2 + tr(\Sigma_1 + \Sigma_2 - 2\sqrt{\Sigma_1 \Sigma_2})
\end{equation}
with $\mu$ the mean of activations, $\Sigma$ the covariance matrix and $tr$ the trace function. The~FID score, as~opposed to pixel-level comparisons, compares the similarity of images at a high level in the feature layers, where significant patterns can be identified. This is in contrast to comparisons that are made at the pixel level. Due to the fact that FID is measured on a collection of images, we are able to make a comparison between the statistics of the distributions found in natural images (or any other set of images that may be desired) and those produced by a generative method. Because~it operates in a manner analogous to that of a distance metric, values that are lower indicate that the generated images are statistically more comparable to either the training or test~data.

Table~\ref{table:1} shows the average FID score results for three different initializations. We can see that across the four datasets, feature alignment has higher scores, which indicates that, in~comparison to the other approaches, it has a lower sampling quality. When a generative network is used, however, feature alignment can achieve scores that are more comparable to those achieved by~GANs.

\begin{table}[h!]
\begin{center}
\begin{tabular}{ | >{\centering\arraybackslash} m{5em} | >{\centering\arraybackslash} m{7em} | >{\centering\arraybackslash} m{7em} | >{\centering\arraybackslash} m{7em} | >{\centering\arraybackslash} m{7em} |}
\hline
Method & MNIST & CelebA & CIFAR-10 & STL-10 \\ 
\hline
VAE  & 39.84 $\pm$ 0.15 & 84.84 $\pm$ 0.10 & 163.59 $\pm$ 0.37 & 201.59 $\pm$ 0.46\\ 
\hline
GAN & 21.50 $\pm$2.79  & 32.85 $\pm$ 1.22 & 63.39 $\pm$ 0.62 & 247.35 $\pm$ 2.43 \\
\hline
$\textbf{VFA}$ (\textbf{ours}) & 120.02 $\pm$ 1.11 & 143.51 $\pm$ 1.04 & 209.32 $\pm$ 1.96 &  259.96 $\pm$ 2.46\\
\hline
$\textbf{VFA-GAN}$ (\textbf{ours}) & 41.24 $\pm$  2.71 & 132.18 $\pm$ 2.73 & 73.20 $\pm$ 2.67 & 167.26 $\pm$ 5.18 \\
\hline
\end{tabular}
\caption{FID scores across four image datasets. Mean and standard deviation from three trials. Lower is better. We can see that the generator improves the quality of the samples by lowering the FID score from sampling variational autoencoders trained with feature alignment.}
\label{table:1}
\end{center}
\end{table}

\unskip

The results of each dataset section below demonstrate that VFA has a lower sample quality and, thus, a greater FID. This behavior is primarily caused by two factors: first, VFA has the same limitations as variational auto-encoders, in which sampling is constrained by the latent layer and tends to produce blurry images to some extent as a result of the loss attempting to approximate the distribution on the latent space to a normal distribution.
Second, in addition to optimizing reconstruction and approaching the normal distribution, the loss of VFA must also optimize reversibility when the encoders of a VAE and VFA are set to have the same size.
The GAN formulation and publications concerning the FID metric both demonstrate that both VFA and VAE contain losses that act on pixel space, which is not always the optimal measure to produce sharp images. Since the generator is trained with a discriminator and both have losses that function on a different space than the pixels, connecting the generator network to enhance the quality of the reconstructed inputs leads to better samples and a lower FID.

\subsection{MNIST}

\begin{figure}[H]
\captionsetup[subfigure]{justification=centering}
    \centering
    \includegraphics[scale=0.75]{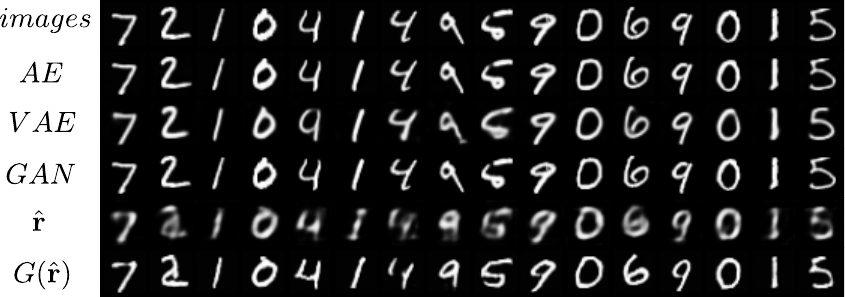}
    \caption{Reconstruction of images of the MNIST dataset from four different~models. }
    \label{fig:mnist_rec}
\end{figure}

\begin{figure}[H]
    \centering
    \includegraphics[scale=0.6]{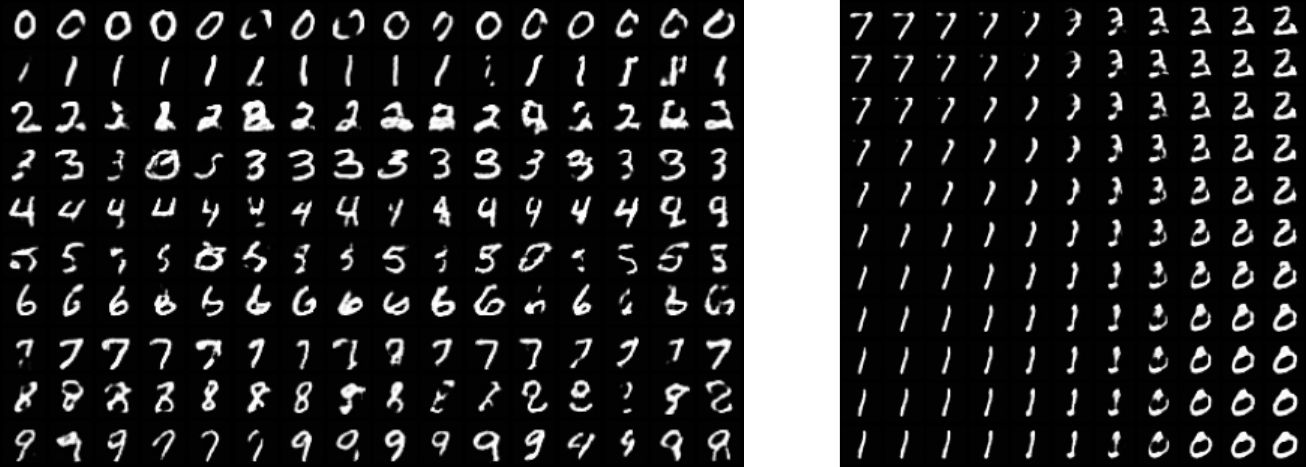}
    \caption{Left: random samples. Right: interpolation among four images reconstructed from the~dataset.}
    \label{fig:sample_mnist}
\end{figure}

The MNIST dataset~\cite{lecun_gradient-based_1998} is a collection of 60,000 grayscale images with size \mbox{$28\times28$ pixels} that contain hand drawings of digits from zero to nine. Figure~\ref{fig:fig5}a shows the latent space trained on two neurons as outputs, we can see that the network attempts to cluster the images to similarity, while Figure~\ref{fig:mnist_latent2} shows the reconstruction for a fixed latent vector but varying a trained classified output vector according to the labels of the dataset. Figure~\ref{fig:mnist_rec} shows the reconstruction of images by the features and with the generator applied to them, compared with traditional AE, VAE, and~GAN. Even though the reconstructions are frequently noisy, the~generator can sharpen the images to make them more similar to the original inputs. Figure~\ref{fig:sample_mnist} shows random samples from the generator with a class layer coupled to the encoder. To~show the consistency of transitions on the latent space, the~same image also features the interpolation of the latent vector between four pairs of~images.

\begin{subfigure}[H]
    \centering
    \setcounter{subfigure}{0}
    \begin{minipage}[b]{0.45\textwidth}
        \includegraphics[width=\linewidth]{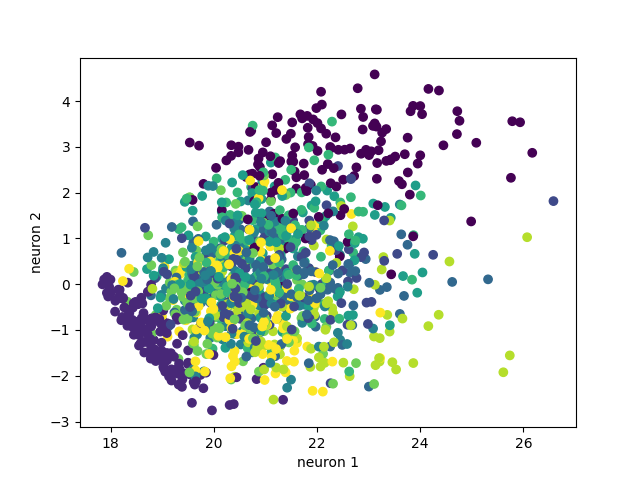}
        \caption{}
        \label{fig:mnist_latent}
    \end{minipage}  
   \setcounter{subfigure}{1}
    \begin{minipage}[b]{0.45\textwidth}
        \includegraphics[width=\linewidth]{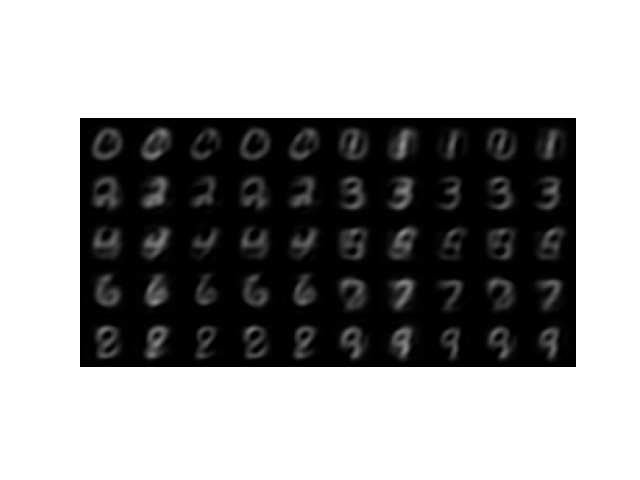}
        \caption{}
        \label{fig:mnist_latent2}
    \end{minipage}
    
    \setcounter{subfigure}{-1}
    \caption{Representation of the latent space. (\textbf{a}) Latent space with two neurons, (\textbf{b}) images from features extracted by manipulating the classification~layer.}
    \label{fig:fig5}
\end{subfigure}

\vspace{-6pt} 

The size of the model and the amount of time required to draw a single sample are provided in the Table~\ref{table:model_size_mnist}. The~size of a feature algorithm is approximately half that of a variational autoencoder because it trains a network without a decoder, but~its execution time is roughly the same due to the backward pass used to update the random~input.

\begin{table}[h!]
\begin{center}
\begin{tabular}{ | >{\centering\arraybackslash} m{5em} | >{\centering\arraybackslash} m{7em} | >{\centering\arraybackslash} m{7em} | >{\centering\arraybackslash} m{7em} | }
\hline
Model & Size ($\times 10^5$ parameters) & Time ($\times 10^{-4} $ s)\\ 
\hline
VAE & 27.40 & 9.44 \\
\hline
GAN & 13.44 & 5.73 \\
\hline
\textbf{VFA} & 14.0 & 11.66 \\
\hline
\textbf{VFA-GAN} & 14.16 & 20.59 \\
\hline
\end{tabular}
\caption{Size and execution time required to draw a single sample using  multiple models trained on the MNIST dataset. VFA takes approximately the same amount of time as VAE, but only half the number of parameters.}
\label{table:model_size_mnist}
\end{center}
\end{table}
\unskip

\subsection{CelebA}

\begin{figure}[H]
\centering
\includegraphics[scale=1.0]{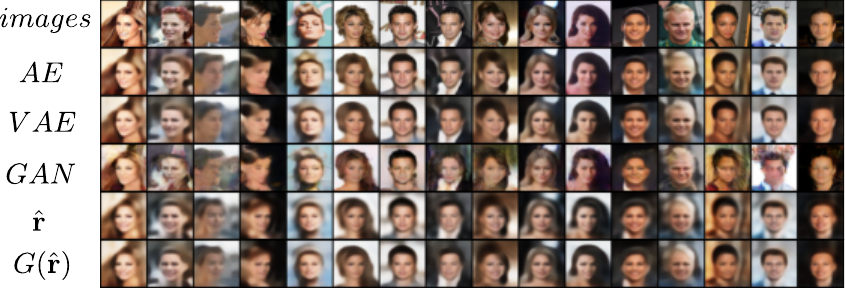}
    \caption{Reconstruction of images of the CelebA dataset from four different~models. }
    \label{fig:celeb_rec}
\end{figure}

\begin{figure}[H]
\centering
 \includegraphics[scale=0.5]{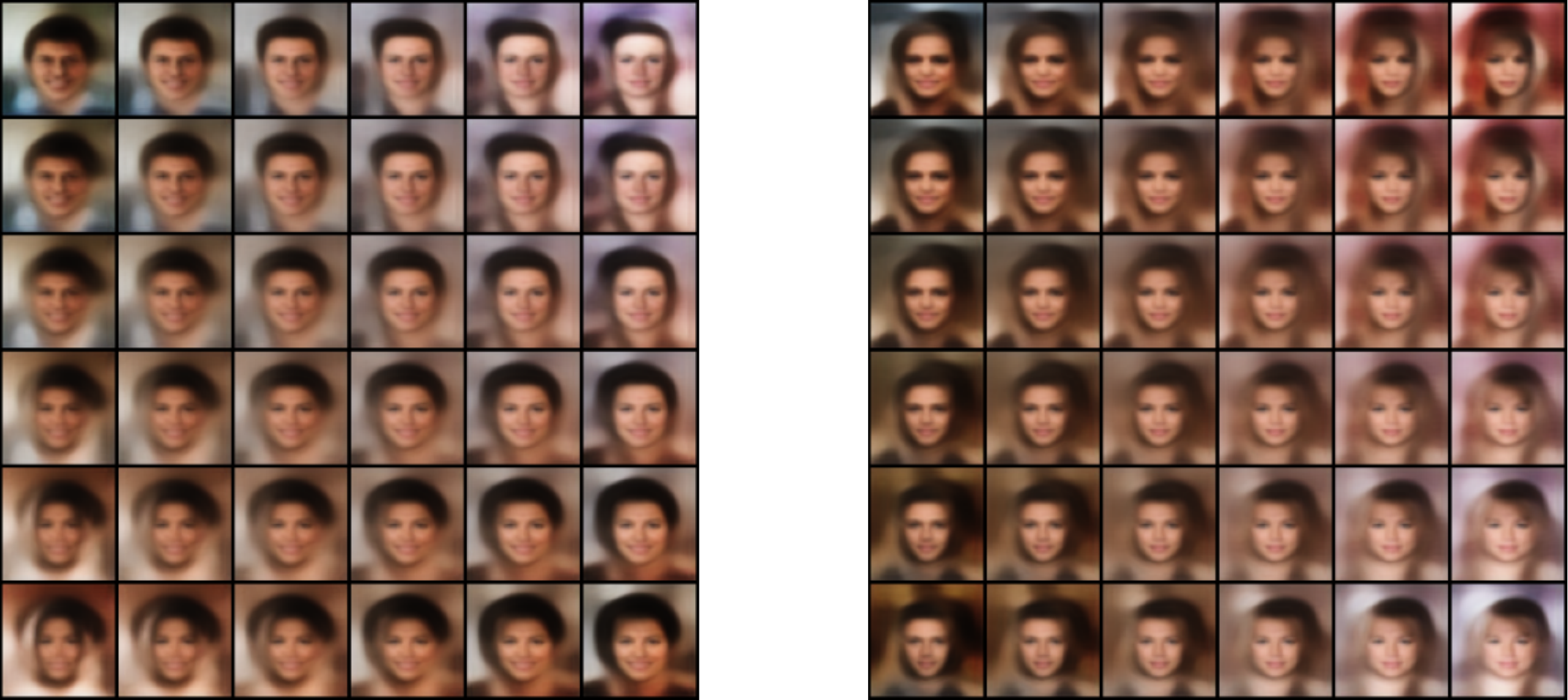}
    \caption{Two sets of interpolation among four random sampled~images.}
    \label{fig:celeb_sample}
\end{figure}

The CelebA dataset~\cite{liu2015faceattributes} is a collection of 202,600 images of celebrity faces. The~images were resized to $64\times64$ pixels. Figures~\ref{fig:celeb_rec} and \ref{fig:celeb_sample} show the reconstruction and sampling with interpolation between samples, respectively. Without~the perceptual loss (with $\lambda=1$), we noticed a failure on the convergence of the generator network, resulting in samples containing only~noise.

\vspace{-6pt} 

The size and duration of sampling a single image for various models are displayed in Table~\ref{table:model_size_celeba}. As~before, we can see that VFA takes about the same amount of time despite being half the~size.

\begin{table}[h!]
\begin{center}
\begin{tabular}{ | >{\centering\arraybackslash} m{5em} | >{\centering\arraybackslash} m{7em} | >{\centering\arraybackslash} m{7em} | >{\centering\arraybackslash} m{7em} | }
\hline
Model & Size ($\times 10^5$ parameters) & Time ($\times 10^{-4} $ s)\\ 
\hline
VAE & 37.48 & 19.41 \\
\hline
GAN & 18.48 & 12.45 \\
\hline
\textbf{VFA} & 19.04 & 22.79 \\
\hline
\textbf{VFA-GAN} & 19.24 & 41.71 \\
\hline
\end{tabular}
\caption{Size and execution time required to draw a single sample using  multiple models trained on the CelebA dataset. VFA takes approximately the same amount of time as VAE, but only half the number of parameters.}
\label{table:model_size_celeba}
\end{center}
\end{table}
\unskip

\subsection{CIFAR-10}

\begin{figure}[H]
    \centering
    \includegraphics[scale=0.75]{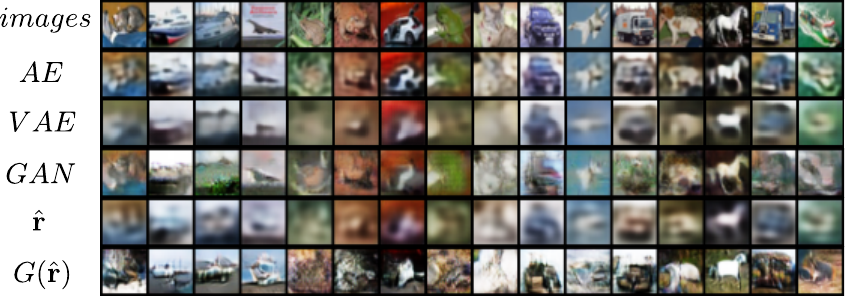}
    \caption{Reconstruction of images on~CIFAR-10.}
    \label{fig:cifar_rec}
\end{figure}

\begin{figure}[H]
\centering
 \includegraphics[scale=0.45]{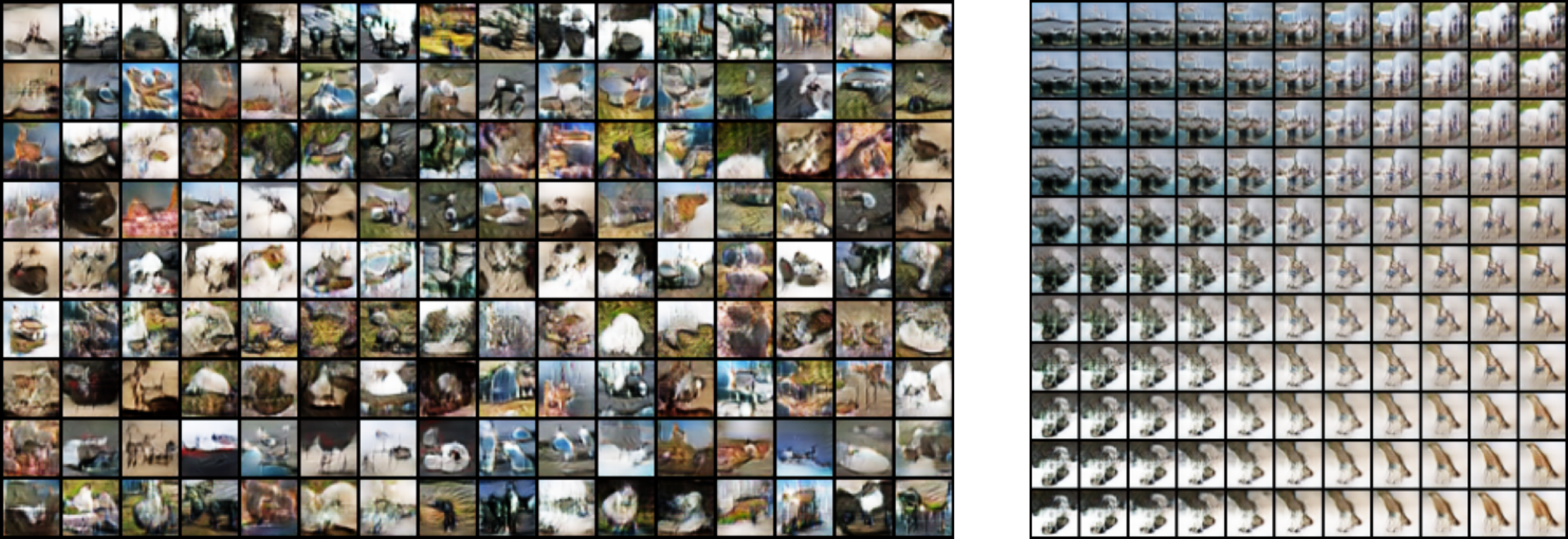}
    \caption{\textbf{Left}: random samples. \textbf{Right}: interpolation among four images reconstructed from the~dataset.}
    \label{fig:cifar_sample}
\end{figure}

The CIFAR-10 dataset \cite{krizhevsky09} contains 70,000 natural images with size $32 \times 32$ pixels across 10 different classes. Figure~\ref{fig:cifar_rec} shows the reconstruction of images from the dataset. While the features do approximate the original inputs, the~transformation of the generator tends to be more dissimilar due to its loss being dependent only on the adversarial contribution ($\lambda=0)$. Just as before, Figure~\ref{fig:cifar_sample} shows random samples and interpolation, which show a diversity of images, albeit less perceptual similar to the original~dataset.

\vspace{-6pt} 

Table~\ref{table:model_size_cifar} shows the size and time of sampling for different models. Since VFA is very similar to VAE, but~without the decoder, we can see that it maintains the half-size pattern for the same amount of time, due to the backward~pass.

\begin{table}[h!]
\begin{center}
\begin{tabular}{ | >{\centering\arraybackslash} m{5em} | >{\centering\arraybackslash} m{7em} | >{\centering\arraybackslash} m{7em} | >{\centering\arraybackslash} m{7em} | }
\hline
Model & Size ($\times 10^5$ parameters) & Time ($\times 10^{-4} $ s)\\ 
\hline
VAE & 97.55 & 10.01 \\
\hline
GAN & 47.48 & 7.20\\
\hline
\textbf{VFA} & 50.28 & 11.25 \\
\hline
\textbf{VFA-GAN} & 51.95 & 21.56 \\
\hline
\end{tabular}
\caption{Size and execution time required to draw a single sample using  multiple models trained on the CIFAR-10 dataset. VFA takes approximately the same amount of time as VAE, but only half the number of parameters.}
\label{table:model_size_cifar}
\end{center}
\end{table}

\subsection{STL-10}

The STL-10 dataset \cite{stldataset} is a subset of the ImageNet dataset that contains 100000 unlabeled images, and a additional of 500 labeled images for training and 800 images for testing. This dataset is mostly used for unsupervised tasks, but since in this work we are interested in image generation, we used only the set of unlabeled images, resized to $64 \times 64$ pixels.

\begin{figure}[H]
    \centering
    \includegraphics[scale=0.3]{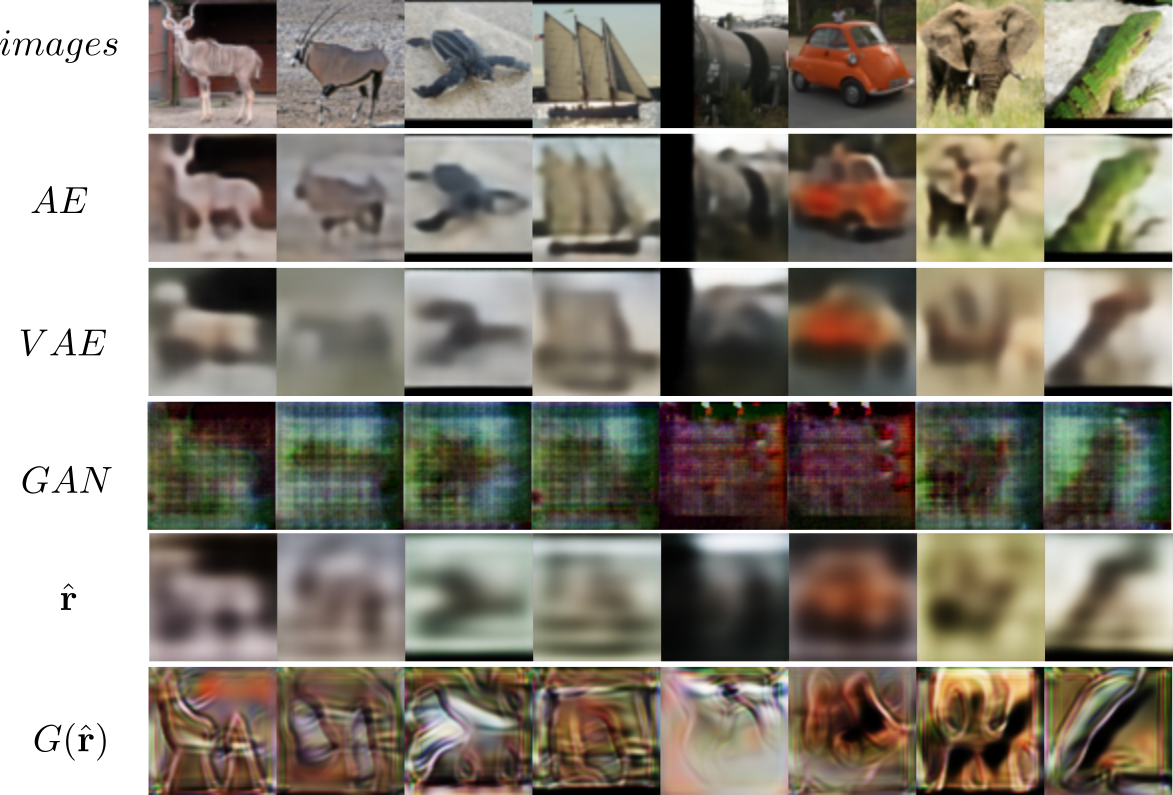}
    \caption{Reconstruction of images on STL-10 from different models.}
    \label{fig:stl_rec}
\end{figure}

The reconstruction results are shown in figure \ref{fig:stl_rec}. We can see that the reconstruction $\hat{r}$ from VFA is visually similar to VAE, but has slightly better fidelity to shapes. 

\vspace{-6pt} 

\begin{figure}[H]
    \centering
    \includegraphics[scale=0.5]{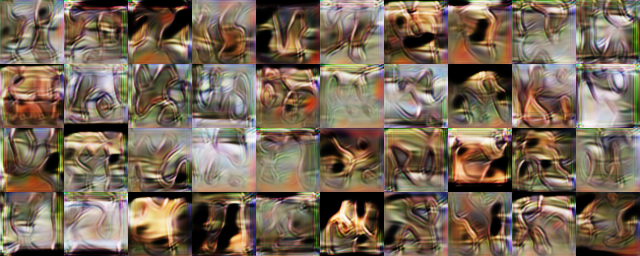}
    \caption{Random samples from VFA-GAN of images on STL-10.}
    \label{fig:stl_sample}
\end{figure}

Random samples from VFA-GAN are shown in Figure \ref{fig:stl_sample}. We also show, in table \ref{table:model_size_stl}, the size of each model and the time required to draw samples. Note that the size of the model \textit{VFA-GAN} includes both the encoder and generator networks.

\vspace{-6pt} 

\begin{table}[h!]
\begin{center}
\begin{tabular}{ | >{\centering\arraybackslash} m{5em} | >{\centering\arraybackslash} m{7em} | >{\centering\arraybackslash} m{7em} | >{\centering\arraybackslash} m{7em} | }
\hline
Model & Size ($\times 10^5$ parameters) & Time ($\times 10^{-2} $ s)\\ 
\hline
VAE & 254.22 & 1.51 \\
\hline
GAN & 32.74 & 1.50\\
\hline
\textbf{VFA} & 221.48 & 1.72 \\
\hline
\textbf{VFA-GAN} & 229.90 & 2.18 \\
\hline
\end{tabular}
\caption{Size and execution time required to draw a single sample using  multiple models trained on the STL-10 dataset.}
\label{table:model_size_stl}
\end{center}
\end{table}

\begin{figure}[]
    \centering
    \includegraphics[scale=0.75]{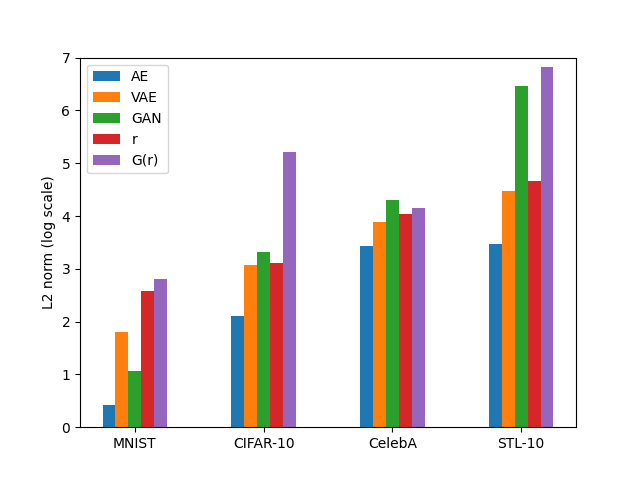}
    \caption{Comparison of the reconstruction $L_{2}$ loss for AE, VAE, GAN, $r$, and~$G(r)$.}
    \label{fig:histogram}
\end{figure}

It is important to note that feature alignment is not expected to outperform the reconstruction and sample qualities of VAEs and GANs.  Because~the reversibility condition is a constraint on neural network optimization, which must thus balance the reversibility cost with other losses. Nevertheless, we compare the reconstruction $L_2$ loss with other networks to analyze how different each network is compared to the same metric. These results are shown in Figure~\ref{fig:histogram}.

Given that it is optimized directly for reconstruction, autoencoders have the lowest loss, which is to be expected given the nature of the network.
Because they are optimized with the same amount of loss, the~reconstructions from feature alignment should be compared to VAEs and the generator should be compared to GAN. It is clear from this that the lack of perceptual loss on the generator network (for the CIFAR-10 dataset) has a negative impact on the reconstruction ability (without first optimizing the latent vector).

\subsection{Local Feature~Alignment}

In this section, we present the results obtained by an encoder that was trained for reconstruction using the local feature alignment training. Figures~\ref{fig:local_mnist}--\ref{fig:local_stl} show reconstruction pairs for  the MNIST, CIFAR-10, CelebA and STL-10 datasets respectively. We can observe that local training can reconstruct images even though the layers do not receive any information from the reconstruction loss of images. This can be attributed to the same reason as non-local feature alignment: the weights form an orthogonal matrix that attempts to reverse information between layers as much as possible, that is only limited by the network capacity, which is directly related to the number of~neurons.

\setcounter{figure}{15}
\begin{subfigure}[H]
    \setcounter{subfigure}{0}
    \centering
    \begin{minipage}[b]{0.75\textwidth}
        \includegraphics[width=\linewidth]{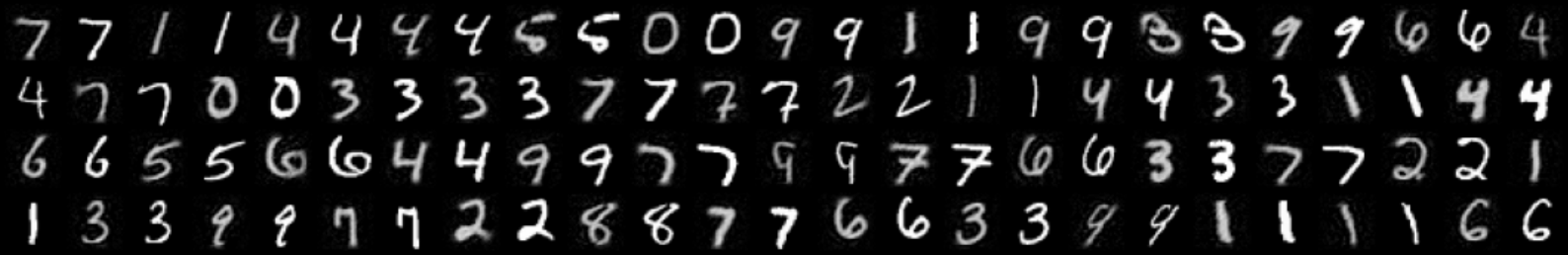}
        \caption{}
        \label{fig:local_mnist}
    \end{minipage}
    
   \setcounter{subfigure}{1}
    \begin{minipage}[b]{0.75\textwidth}
        \includegraphics[width=\linewidth]{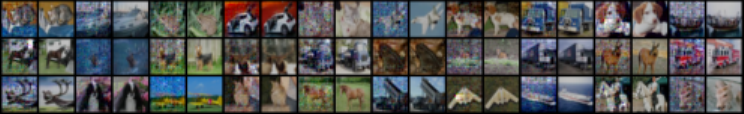}
        \caption{}
        \label{fig:local_cifar}
    \end{minipage}
    
    \setcounter{subfigure}{2}
    \begin{minipage}[b]{0.75\textwidth}
        \includegraphics[width=\linewidth]{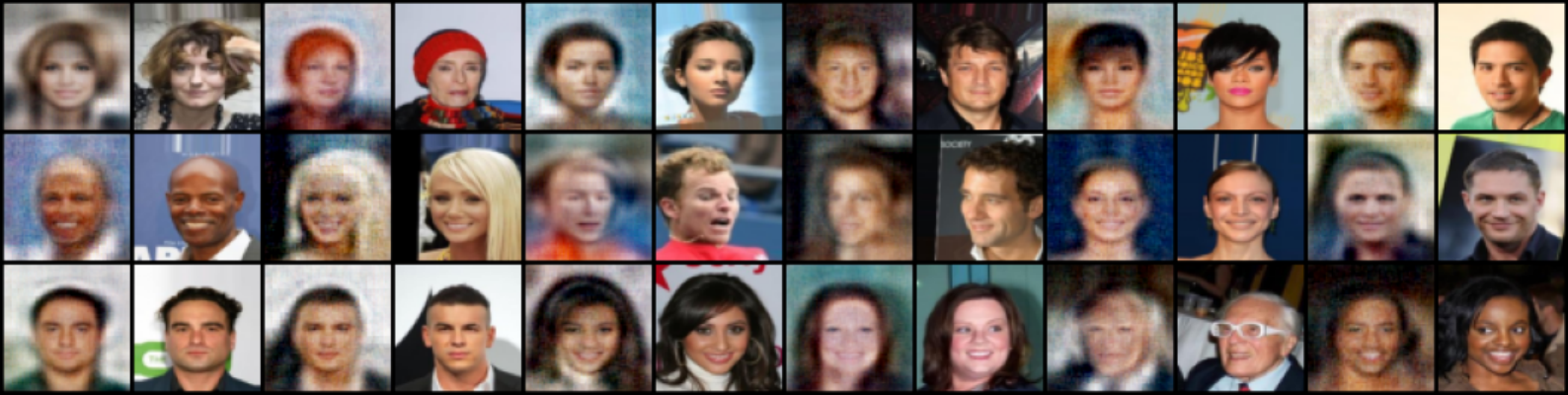}
        \caption{}
        \label{fig:local_celeb}
    \end{minipage}
    
    \setcounter{subfigure}{3}
    \begin{minipage}[b]{0.75\textwidth}
        \includegraphics[width=\linewidth]{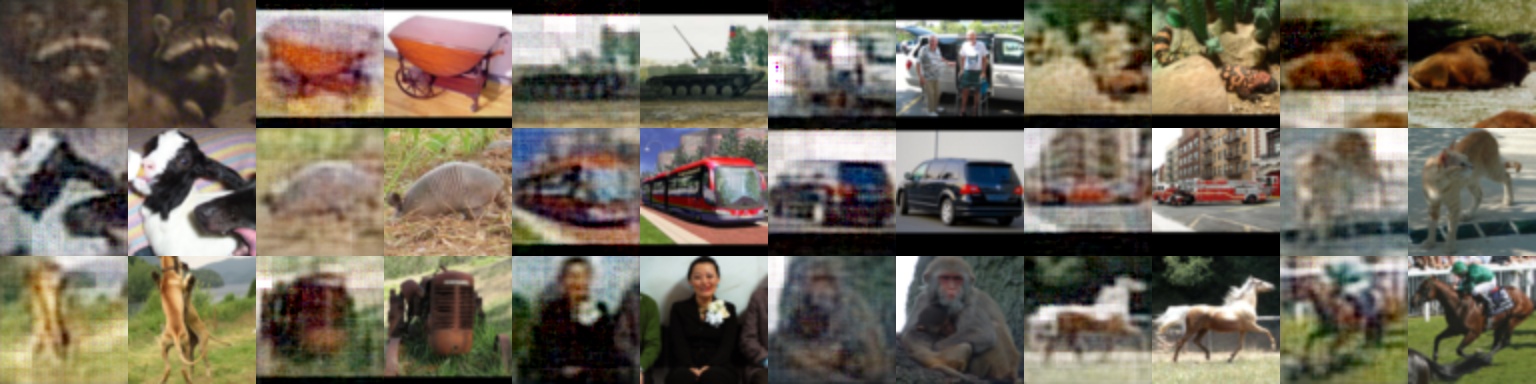}
        \caption{}
        \label{fig:local_stl}
    \end{minipage}
    
    \setcounter{subfigure}{-1}
    \caption{Local feature alignment. Each pair of images contains the reconstruction and original, respectively. \textbf{(a)} MNIST, \textbf{(b)} CIFAR-10, \textbf{(c)} CelebA, \textbf{(d)} STL-10.}
    \label{fig:local_subfigures}
\end{subfigure}

\section{Conclusions}

We presented feature alignment, a technique to approximate reversibility in neural networks. By optimizing the features to match the inputs, we trained an encoder to predict its input, given an output. For a simple case, we showed that it is possible to recover the inputs given only the outputs by adding latent variables, which are optimized only with the reversibility loss. We can generate new samples with the same statistical distribution as the training data by coupling a probabilistic layer with the same formulation as the variational autoencoders. We combined the generative adversarial network method by coupling a generator and a discriminator network to the images generated by the method to improve the quality of the generated samples, which suffer from noise effects. We also demonstrated that the technique can be modified to use a local training rule instead of backpropagation, which has the advantage of using less memory for training and extracting gradients from neural~networks.

Mathematical analysis on the convergence of the proposed technique shows that the weights converge to a pseudo-inverse matrix, which justifies the convergence of a network trained in this way to map its outputs back to its inputs. Since the bottlenecks do not permit a one-to-one relationship, the~restriction is the architecture of the network~itself.

We used the technique to reconstruct and generate images from the datasets MNIST, CIFAR-10, CelebA and STL-10. The~results demonstrate that the features can approximate the inputs. Despite the fact that it cannot improve on the sampling quality of other current generative techniques, reversibility can be advantageous when a mapping of the outputs back to their inputs is desired. The CIFAR-10 and STL-10 datasets are notoriously difficult due to the small image size and high variance, resulting in samples with a high FID measure. 

The primary shortcoming of the approach is that it places a restriction of reversibility on the parameters of a neural network. This forces the parameters to be balanced between reconstruction or sampling and reversibility, which in turn reduces the quality of the images that are produced. When an encoder is trained to compress an image into a latent representation, which must approximate a normal distribution, and also reconstruct the input given this representation, this can be seen as an additional loss to optimize. An encoder is trained to accomplish these two tasks simultaneously.
Therefore, in order to recover the inputs, the trainable parameters of a network not only need to minimize some loss with respect to the forward propagation of information, but they also need to reduce some loss with respect to the backward propagation of information.
As a direct consequence of this, there is a limited number of practicable optimal configurations that the trainable parameters are capable of reaching.

We can use label information for conditional sample generation by connecting a classification layer to the encoder network.
Furthermore, the~results of local training suggest that we can train neural networks without using global loss function feedback, which is an important area of application of this technique.

The proposed method can be interpreted in a way that places it somewhere in the middle of VAEs and GANs.
The complete architecture, which is comprised of an encoder and a generator network, possesses more complex latent vectors that may be exploited and generates samples that are crisper than those produced by VAEs. Feature alignment is a technique that can be implemented across a broad variety of neural network designs so long as the architecture of the neural networks being utilized can be entirely differentiated end to end. As a result, it offers the possibility of making modifications, which could potentially lead to an improvement in the results in general. For instance, utilizing residual networks, such as ResNets \cite{resnet} for the encoder and U-Net \cite{unet} for the generator network, since the latter can transfer the dimension of the input onto itself, are two examples of how this may be done to both improve the flow of information and reduce the influence of disappear gradients.

\section{Appendix A: Mathematical analysis}

This additional material studies the numerical analysis of the~technique.

\subsection{Convergence of the~Features}\label{A.1}

For two fully connected layers, we have $a_j^{(x)} = \hat{a}_j = \sum_i w_{ij} x_i$ and $a_j^{(r)} = \sum_i w_{ij} r_i$. The~feature is obtained by optimizing the squared $L_2$ loss between the two activations:
\begin{align} 
\mathcal{L} = \frac{1}{2} (\hat{a}_j - a_j^{(r)})^2.
\end{align}

It follows that $r$ will evolve with the gradient flux:
\begin{align} 
\frac{\partial r_i}{\partial t} = -\frac{\partial \mathcal{L}}{\partial r_i} \ \therefore
r_i^t = r_i^{t-1} - \frac{\partial C}{\partial r^{t-1}}.
\end{align}

So, we can evaluate $r^t$ at each time step $t$:
\begin{align} 
& r_i^1 = r_i^{0} - w_{ij}(\hat{a}_j - w_{ij}r^0)= r_i^{0}(1- w_{ij}^2) + w_{ij}\hat{a}_j, \\
& r_i^2 = r_i^{1}(1- w_{ij}^2) + w_{ij}\hat{a}_j = r_i^{0}(1- w_{ij}^2)^2 + [1 + (1 - w_{ij}^2)]w_{ij}\hat{a}_j, \\
& r_i^3 = r_i^{0}(1- w_{ij}^2)^3 + [1 + (1 - w_{ij}^2) + (1-w_{ij}^2)^2]w_{ij}\hat{a}_j, \\
& r_i^4 = r_i^{0}(1- w_{ij}^2)^4 + [1 + (1 - w_{ij}^2) + (1-w_{ij}^2)^2 + (1-w_{ij}^2)^3]w_{ij}\hat{a}_j, \\
& r_i^5 = ...
\end{align}

From the pattern above, we generalize $r^t$ for any time step as follows:
\begin{align} 
r_i^p = r_i^{0}(1- w_{ij}^2)^p + \sum_{q=0}^{p-1}(1-w^2)^q w_{ij}\hat{a}_j.
\end{align}

Under the restriction of $w_{ij}^2 \le 2$, as~$t$ grows we have as a limit case:
\begin{align} 
\lim_{p\rightarrow \infty} r_i^{0}(1- w_{ij}^2)^p = 0 \ \therefore 
\lim_{p\rightarrow \infty} \sum_{q=0}^{p-1}(1-w_{ij}^2)^q w_{ij}\hat{a}_j = \frac{\hat{a}_j}{w_{ij}}.
\end{align}

So the loss $\mathcal{L} \rightarrow 0$ as $p\rightarrow \infty$.

\subsection{Convergence of the~Weights}\label{A.2}
The $L_2$ loss, which updates the parameters, is:
\begin{align} 
\mathcal{L} = \frac{1}{2}(x_i - r_i^t)^2 = \frac{1}{2}\left[x_i - r_i^{0}(1- w_{ij}^2)^t - \sum_{q=0}^{t-1}(1-w_{ij}^2)^q w_{ij}\hat{a}_j\right]^2.
\end{align}

For one-shot, $T=1$, training, we have:
\begin{align} 
\mathcal{L} = \frac{1}{2}\left[x_i - r_i^{0}(1- w_{ij}^2) - w_{ij}\hat{a}_j\right]^2 = \frac{1}{2}\left[x_i - r_i^{0}(1- w_{ij}^2) - w_{ij}w_{ij}x_i\right]^2.
\end{align}

We can rewrite the equation above in vector notation as follows:
\begin{align} 
    \mathcal{L} = \frac{1}{2}\left[\textbf{x} - \textbf{r}^{0}(\textbf{I} - \textbf{w}^T \textbf{w}) - \textbf{w}^T \textbf{w} \textbf{x}\right]^2.
\end{align}

For any $\textbf{r}^0$, the~equation above has roots for $w_{ii}^2=0$.
We can see then that the loss is minimal when the weight matrix product is orthogonal, i.e.,~$\textbf{w}^T \textbf{w} = \textbf{I}$. This has as a consequence that the transposed weight matrix is also its generalized Moore–Penrose inverse or pseudo-inverse $\textbf{w}^{T} = \textbf{w}^{-1}$.

\section{Appendix B: List of networks}\label{B.1}

This section lists the networks used for each dataset for feature alignment. The~notation $Conv2d(f, k, s, p)$ means output filters $f$, kernel size $kxk$, stride $s$ and padding $p$, while $Linear(n)$ has $n$ fully connected~neurons.

\begin{table}[H]
\begin{center}
\begin{tabular}[t]{cc}
  \hline
  Input \boldmath{$1 \times~28 \times~28$} \\ 
  \hline
  Conv2d(32, 3, 1, 1) + LeakyReLU \\ 
  \hline
  Conv2d(32, 3, 2, 1) + LeakyReLU \\ 
  \hline
  Conv2d(64, 3, 1, 1) + LeakyReLU \\ 
  \hline
  Conv2d(64, 3, 2, 1) + LeakyReLU + Flatten\\ 
  \hline
  Linear(4096) + LeakyReLU\\
  \hline
  Linear(Z)\\
  \hline
\end{tabular}
\end{center}
\caption{\label{tab:network_mnist}Encoder for MNIST.}
\end{table}

\begin{table}[H]
\begin{center}
\begin{tabular}[t]{cc}
  \hline
  Input \boldmath{$3\times~32 \times~32$} \\ 
  \hline
  Conv2d(32, 3, 1, 1) + LeakyReLU \\ 
  \hline
  Conv2d(32, 3, 2, 1) + LeakyReLU \\ 
  \hline
  Conv2d(64, 3, 1, 1) + LeakyReLU \\ 
  \hline
  Conv2d(64, 3, 2, 1) + LeakyReLU \\ 
  \hline
  Conv2d(128, 3, 1, 1) + LeakyReLU \\ 
  \hline
  Conv2d(128, 3, 2, 1) + LeakyReLU + Flatten\\ 
  \hline
  Linear(2048) + LeakyReLU\\
  \hline
  Linear(Z)\\
  \hline
\end{tabular}
\end{center}
\caption{\label{tab:network_cifar10}Encoder for CIFAR-10.}
\end{table}

\begin{table}[H]
\begin{center}
\begin{tabular}[t]{cc}
  \hline
  Input \boldmath{$3\times~64 \times~64$} \\ 
  \hline
  Conv2d(32, 3, 1, 1) + LeakyReLU \\ 
  \hline
  Conv2d(32, 3, 2, 1) + LeakyReLU \\ 
  \hline
  Conv2d(64, 3, 1, 1) + LeakyReLU \\ 
  \hline
  Conv2d(64, 3, 2, 1) + LeakyReLU\\ 
  \hline
  Conv2d(128, 3, 1, 1) + LeakyReLU \\ 
  \hline
  Conv2d(128, 3, 2, 1) + LeakyReLU \\ 
  \hline
  Conv2d(256, 3, 1, 1) + LeakyReLU \\ 
  \hline
  Conv2d(256, 3, 2, 1) + LeakyReLU + Flatten\\ 
  \hline
  Linear(4096) + LeakyReLU\\
  \hline
  Linear(Z)\\
  \hline
\end{tabular}
\end{center}
\caption{\label{tab:network_celeba}Encoder for CelebA and STL-10. The models for MNIST and CIFAR-10 follow the same pattern, but without the last two convolutions.}
\end{table}

\begin{table}[H]
\begin{center}
\begin{tabular}[t]{cc}
  \hline
  Input \boldmath{$3\times~64 \times~64$} \\ 
  \hline
  Conv2d(32, 3, 1, 1) + LeakyReLU \\ 
  \hline
  Conv2d(32, 3, 2, 1) + LeakyReLU + BatchNorm2d(32)\\ 
  \hline
  Conv2d(64, 3, 1, 1) + LeakyReLU \\ 
  \hline
  Conv2d(64, 3, 2, 1) + LeakyReLU + BatchNorm2d(64)\\ 
  \hline
  Conv2d(128, 3, 1, 1) + LeakyReLU \\ 
  \hline
  Conv2d(128, 3, 2, 1) + LeakyReLU + BatchNorm2d(128)\\
  \hline
  Conv2d(256, 3, 1, 1) + LeakyReLU \\ 
  \hline
  Conv2d(256, 3, 2, 1) + LeakyReLU + + BatchNorm2d(256) + Flatten\\ 
  \hline
  Linear(4096) + LeakyReLU\\
  \hline
  Linear(1)\\
  \hline
\end{tabular}
\end{center}
\caption{\label{tab:network_discriminator} Discriminator network for CelebA and STL-10. The models for MNIST and CIFAR-10 follow the same pattern, but without the last two convolutions.}
\end{table}

\begin{table}[H]
\begin{center}
\begin{tabular}[t]{cc}
  \hline
  Input \boldmath{$3\times~64 \times~64$} \\ 
  \hline
  Conv2d(128, 7, 1, 1) + LeakyReLU \\ 
  \hline
  Conv2d(128, 7, 1, 1) + LeakyReLU \\ 
  \hline
  Conv2d(3, 7, 1, 1) + Sigmoid \\ 
  \hline
\end{tabular}
\end{center}
\caption{\label{tab:network_generator} Generator network for CelebA and STL-10 used with the feature alignment.}
\end{table}

\begin{table}[H]
\begin{center}
\begin{tabular}[t]{cc}
  \hline
  Input \boldmath{$3\times~64 \times~64$} \\ 
  \hline
  Linear(Z, 4096) + LeakyReLU + reshape((256,4,4)) \\
  \hline
  ConvTranspose2d(256, 3, 2, 1) + LeakyReLU \\ 
  \hline
  ConvTranspose2d(128, 3, 1, 1) + LeakyReLU \\ 
  \hline
  ConvTranspose2d(128, 3, 2, 1) + LeakyReLU \\ 
  \hline
  ConvTranspose2d(64, 3, 1, 1) + LeakyReLU \\ 
  \hline
  ConvTranspose2d(64, 3, 2, 1) + LeakyReLU \\
  \hline
  ConvTranspose2d(32, 3, 1, 1) + LeakyReLU \\ 
  \hline
  ConvTranspose2d(32, 3, 2, 1) + LeakyReLU \\ 
  \hline
  Conv2d(3, 3, 1, 1) + Sigmoid \\ 
\end{tabular}
\end{center}
\caption{\label{tab:network_gan} Generator network for CelebA and STL-10 used with the GAN method. The models for MNIST and CIFAR-10 follow the same pattern, but without the last two convolutions.}
\end{table}

\section*{Conflict of Interest Statement}

The authors declare that the research was conducted in the absence of any commercial or financial relationships that could be construed as a potential conflict of interest.
\section*{Funding}
This work was supported by the  National Institute for the Science and Technology of Quantum Information (INCT-IQ), process 465469/2014-0, and by the National Council for Scientific and Technological Development (CNPq), processes 309862/2021-3 and 140758/2019-4.

\section*{Data Availability Statement}
The code to reproduce the results is available on Github: \url{https://github.com/tiago939/feature_alignment} (accessed on 20 August 2022.

\bibliographystyle{Frontiers-Harvard} 
\bibliography{frontiers}

\end{document}